\documentclass[conference]{IEEEtran}
\IEEEoverridecommandlockouts
\usepackage{cite}
\usepackage{amsmath,amssymb,amsfonts}
\usepackage{algorithmic}
\usepackage{graphicx}
\usepackage{textcomp}
\usepackage{xcolor}
\usepackage{multirow}
\usepackage{booktabs}
\usepackage{rotating}
\usepackage{xcolor}
\usepackage{xcolor,colortbl}
\usepackage{soul}
\usepackage{caption}
\def\BibTeX{{\rm B\kern-.05em{\sc i\kern-.025em b}\kern-.08em
    T\kern-.1667em\lower.7ex\hbox{E}\kern-.125emX}}
\begin{document}



\title{BioALBERT: A Simple and Effective \\Pre-trained Language Model for \\Biomedical Named Entity Recognition
}

\author{\IEEEauthorblockN{Usman Naseem\IEEEauthorrefmark{1},
Matloob Khushi\IEEEauthorrefmark{2}
Vinay Reddy\IEEEauthorrefmark{3},
Sakthivel Rajendran\IEEEauthorrefmark{4},
Imran Razzak\IEEEauthorrefmark{5} and
Jinman Kim\IEEEauthorrefmark{6}}

\IEEEauthorblockA{\IEEEauthorrefmark{1}\IEEEauthorrefmark{2}\IEEEauthorrefmark{3}\IEEEauthorrefmark{5}\IEEEauthorrefmark{6}School of Computer Science,
University of Sydney, Australia\\
\IEEEauthorrefmark{4}School of Information Technology, Deakin University, Australia\\
Email:\IEEEauthorrefmark{1}\IEEEauthorrefmark{2}\IEEEauthorrefmark{3}\IEEEauthorrefmark{4}\IEEEauthorrefmark{6}FirstName.SecondName@sydney.edu.au,\IEEEauthorrefmark{5}FirstName.SecondName@deakin.edu.au}}



\maketitle

\begin{abstract}

In recent years, with the growing amount of biomedical documents, coupled with advancement in natural language processing algorithms, the research on biomedical named entity recognition (BioNER) has increased exponentially. However, BioNER research is challenging as NER in the biomedical domain are: (i) often restricted due to limited amount of training data, (ii) an entity can refer to multiple types and concepts depending on its context and, (iii) heavy reliance on acronyms that are sub-domain specific. Existing BioNER approaches often neglect these issues and directly adopt the state-of-the-art (SOTA) models trained in general corpora which often yields unsatisfactory results. We propose biomedical ALBERT (A Lite Bidirectional Encoder Representations from Transformers for Biomedical Text Mining) – bioALBERT – an effective domain-specific language model trained on large-scale biomedical corpora designed to capture biomedical context-dependent NER. We adopted a self-supervised loss used in ALBERT that focuses on modelling inter-sentence coherence to better learn context-dependent representations and incorporated parameter reduction techniques to lower memory consumption and increase the training speed in BioNER. In our experiments, BioALBERT outperformed comparative SOTA BioNER models on eight biomedical NER benchmark datasets with four different entity types. We trained four different variants of BioALBERT models which are available for the research community to be used in future research.

\end{abstract}


\section{Introduction}


The growing volume of the published biomedical literature, such as clinical reports~\cite{meystre2008extracting} and health literacy~\cite{maartensson2012health}, are fuelling the advancements in the development of text mining algorithms. Biomedical named entity recognition (BioNER) intends to automatically identify biomedical entities such as diseases, chemicals, genes and proteins, etc., from the biomedical literature. So, a crucial step towards this aim is to build better and effective methods which can automatically recognize and extract biomedical entities. BioNER is an essential building block of many downstream text mining applications such as extracting drug-to-drug interactions~\cite{vvarticle} and disease-treatment~\cite{rosario-hearst-2004-classifying} relationships. Traditionally, BioNER relies on feature engineering methods (e.g., lexicon-based, rules-based and statistics-based). However, feature engineering is dependent on domain-specific knowledge which does not perform well on BioNER~\cite{cohen2005survey}.

Deep learning (DL) with its ability to automatically extract features have become common in BioNER recently~\cite{yadav2019survey}. For instance, Long Short-Term Memory (LSTM) is usually employed to learn vector representations of each word in a sentence, and then as the input to conditional random fields (CRF)  have greatly improved the performance in BioNER~\cite{yadav2019survey}. Recently state-of-the-art (SOTA) DL based language models such as Embeddings from Language Models (ELMo)~\cite{peter}, Bidirectional Encoder Representations from Transformers (BERT)~\cite{devlin-etal-2019-bert} and (A Lite Bidirectional Encoder
Representations from Transformers (ALBERT)~\cite{lan2019albert} obtained SOTA best performance on many NLP tasks.


Although these models show promising results, but applying them on BioNER has multiple challenges and limitations: (i) a limited amount of training data; (ii) an entity could represent multiple entity types depending on its textual context, i.e., a “\textit{BRCA1}” can be referred as gene name as well as a  disease entity depending on its context. Similarly, “\textit{heart attack}” and “\textit{myocardial infarction}” refer to the same concept and, (iii) the heavy use of acronyms in biomedical texts makes it challenging to identify concepts, i.e., the abbreviation ”\textit{RA}” may refer to “\textit{right atrium},” “\textit{rheumatoid arthritis},” or one of several other concepts, where the resolution of the abbreviation is, therefore, context-dependent.  Therefore, current models in BioNER rely on various context-independent and transformer-based context-dependent language models which are trained on biomedical corpora~\cite{Pyysalo2013DistributionalSR, jin2019probing,lee2019biobert}.

\begin{figure*}[!htpb]
\centering
\includegraphics[width=1\textwidth]{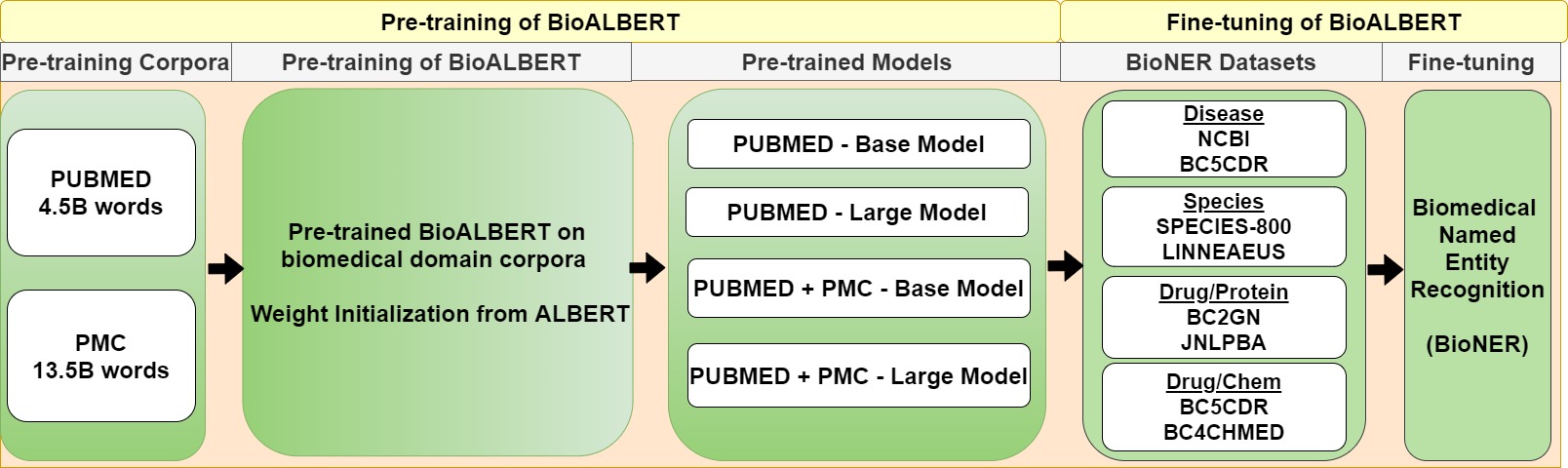}
\caption{ Overview of the pre-training and fine-tuning of BioALBERT on NER}
\label{model}
\end{figure*}


To overcome the identified limitations, we present Biomedical ALBERT – bioALBERT - a context-dependent, fast and effective language model that addresses the shortcomings of recently proposed domain-specific language models. BioALBERT is trained on large biomedical corpora which address the limitation of limited training data.  We also innovate in the adoption of cross-layer parameter sharing by learning parameters for the first block and reuse the block in the remaining layers instead of learning unique parameters for each of the layers and  sentence-order-prediction (SOP) technique as a measure of coherence loss between sentences. SOP takes two consecutive sentences from training data and creates a random pair from different sentences which helps the model to learn better representations and finally, in BERT based models the size of the embedding was linked to the hidden layer sizes of the transformer blocks. These embeddings are projected directly to the hidden space of the hidden layer whereas in our model we use factorized embedding parameterization which decomposes embedding matrix into two small matrices which separate the size of the hidden layers from the size of vocabulary embeddings. This allows for increasing the hidden size without significantly increasing the parameter size of the vocabulary embeddings. BioALBERT is simple and efficient when fine-tuned for BioNER task as compared to other SOTA models.  We evaluate our model on eight biomedical named entity recognition benchmark datasets. Our pre-trained BioNER models, along with the source code, will be publicly available.

\section{Related Work} \label{rw}



\subsection{Language Model}

In biomedical text mining research, there is a long history of using shared language representations to capture the semantics of the text. One established trend is a form of word-embeddings~\cite{mikolov2013distributed} that represent syntactic and semantic meaning, and map words into low-dimensional vectors. Similar methods also have been derived for improving embeddings of word sequences by introducing sentence embeddings~\cite{Chen2018BioSentVecCS}. These context-independent word embeddings approach such as word2vec~\cite{mikolov2013distributed} were trained on biomedical corpora that contain terms and expressions that are usually not included in a general domain corpus~\cite{Pyysalo2013DistributionalSR}. These methods always require complex neural networks to be effectively used and model context-independent representations.

Another common trend, particularly in recent years, is text representation based on context~\cite{peter,devlin-etal-2019-bert,lan2019albert}. Unlike traditional word-embeddings, this enables a word to change its meaning depending on the context in which it occurs. Several other works have investigated the usefulness of contextual models in clinical and biomedical domains. Several researchers trained ELMo on biomedical corpora and presented BioELMo and found that BioELMo beats ELMo on BioNER tasks~\cite{jin2019probing, Zhu_Paschalidis_Tahmasebi_2018}. A pre-trained BioELMo model was released along with their work, allowing more clinical research in the area of BioNER. Beltagy et al.~\cite{beltagy2019scibert} released Scientific BERT (SciBERT), where BERT  was trained on the scientific texts. BERT has typically been superior and better than ELMo to non-contextual embeddings.

Si et al.~\cite{Si_2019}, trained the BERT on clinical notes corpora, using complex task-specific models to improve both traditional embedding and ELMo embedding on the i2b2 2010 and 2012 BioNER. Similarly in another study, a new domain-specific language model, BioBERT~\cite{lee2019biobert}, trained a BERT model on a corpus of biomedical articles from PMC abstracts\footnote{https://www.ncbi.nlm.nih.gov/pmc/}  as well as full texts sourced from PubMed\footnote{https://www.ncbi.nlm.nih.gov/pubmed/}  which gave rise to enhanced performance on BioNER. Peng et al~\cite{peng2019transfer} introduced Biomedical Language Understanding Evaluation (BLUE), a collection of resources for evaluating and analysing biomedical natural language representation models. Their study also confirmed that the BERT models pre-trained on PubMed abstracts and clinical notes see better performance than most state-of-the-art models. 



Despite this success, BERT has some limitations such as BERT has a huge number of parameters which is the cause for problems like degraded pre-training time, memory management issues and model degradation etc~\cite{lan2019albert}. These issues are very well addressed in ALBERT, which is modified based on the architecture of BERT and proposed by Lan et al.~\cite{lan2019albert}. ALBERT incorporates two-parameter reduction techniques that lift the significant obstacles in scaling pre-trained models. (i) Factorized embedding parameterization - decomposes the large vocabulary embedding matrix into two small matrices, (ii) replaces the NSP loss by SOP loss; and  (iii) Cross-layer parameter sharing- prevents the parameter from growing with the depth of the network. These techniques significantly reduce the number of parameters used when compared with BERT without significantly affecting the performance of the model, thus improving parameter-efficiency. An ALBERT configuration similar to BERT-large has 18x fewer parameters and can be trained about 1.7x faster.




\subsection{Fine tuning}

Fine-tuning has been successfully used to transfer pre-trained weights as initialisation for parameters for various downstream tasks~\cite{howard}. This improves the efficiency of the target task when we have limited data and similar tasks~\cite{yogatama2019learning}. Recently, fine-tuning of pre-trained language models have been widely used in various text mining tasks~\cite{qiu2020pre}. Liu et al.~\cite{liu2016recurrent} pre-trained LSTM with a language model and fine-tuned it, and this has contributed to improved performance for various text classification tasks.

Universal Language Model Fine-tuning (ULMFiT) used general-domain pre-training and fine-tuning techniques to avoid over-fitting and achieved SOTA performance on the datasets with less samples~\cite{howard}. Similarly, authors of BERT, ALBERT, BioBERT and others tested the performance of their model for a wide range of tasks with minimum fine-tuning efforts and achieved good performance. Recently, multi-task fine-tuning has led to improvements even with many target tasks~\cite{liu2019multi}. In this paper, we fine-tuned our pre-trained language model BioALBERT, which is trained on biomedical corpora for BioNER. BioALBERT offers better performance on BioNER as it addresses the shortcomings of BERT used in BioBERT. However, this is not a trivial task as it requires several optimisations which are discussed in the next section. Weights of our pre-trained model along with complete source code, will be available online.

\begin{table}[!tpb]
\centering
\caption{Summary of parameters used in the Pre-training}
\label{parameters}
\begin{tabular}{|c|c|c|}
\hline
\multicolumn{3}{|c|}{\textbf{Summary of All parameters used: (Pre-Training)}} \\ \hline
\textbf{Name} & \textbf{BioALBERT 1.0} & \textbf{BioALBERT 1.1} \\ \hline
Architecture & ALBERT Base & ALBERT Large \\ \hline
Activation function & GeLU & GeLU \\ \hline
Number of Attention Heads & 12 & 16 \\ \hline
Number of layers & 12 & 24 \\ \hline
Hidden Size & 768 & 1024 \\ \hline
Embedding Size & 128 & 128 \\ \hline
Vocab Size & 30000 & 30000 \\ \hline
Optimizer & LAMB & LAMB \\ \hline
Train Batch   Size & 1024 & 256 \\ \hline
Eval Batch   Size & 16 & 16 \\ \hline
Max Seq Length & 512 & 512 \\ \hline
Max   Predictions per Seq & 20 & 20 \\ \hline
Learning Rate & 0.00176 & 0.00062 \\ \hline
Training Steps   (PubMed) & 200K & 200K \\ \hline
Training Steps   (PMC) & 270K & 270K \\ \hline
Warmup   Steps & 3125 & 3125 \\ \hline
\end{tabular}%

\end{table}

\section{Methodology} \label{aaamodel}

First, we initialized BioALBERT with weights from ALBERT. BioALBERT is pre-trained on biomedical domain corpora (PubMed abstracts and PMC full-text articles). To show the effectiveness of our approach, BioALBERT is fine-tuned on BioNER. We tested various pre-training strategies with different combinations and sizes of biomedical corpora. The overview of our methodology is shown in Fig.~\ref{model}. Below we present details of each step involved in pre-training and fine-tuning of BioALBERT.

\subsection{Pre-training of BioALBERT}

In this section, we present steps involved in the pre-training of BioALBERT, which has the same architecture as ALBERT, which makes it simple. BioALBERT is trained on PubMed abstracts and PMC full-text articles which contains biomedical terms and enables to pre-train the ALBERT model, which was pre-trained on the general text on biomedical corpora. As we cannot use the raw text biomedical corpora (PubMed and PMC) to pre-train the model, so we converted raw text to structured format by converting raw text files into a single sentence in which; (i) all blank lines are removed within a document and converted it into on single paragraph; (ii) Any line less than 20-character length is removed and (iii) there will be a blank line between each document (when combining the multiple files) for training the model. Overall, PubMed contains approximately 4.5 Billion words, and PMC contains about 13.5 Billion words. As an initial step, we have initialized the model weights from ALBERT to create BioALBERT model by pre-training on biomedical corpora and kept the original vocabulary of ALBERT for pre-training.



\begin{table}[!tpb]
\centering
\caption{BioALBERT Pre-trained Models}
\label{versions}
\resizebox{.50\textwidth}{!}{%
\begin{tabular}{|c|c|c|c|c|c|c|}
\hline
\textbf{Model Version} & \textbf{Model} & \textbf{Trained On} & \textbf{\# of words} & \textbf{Machine Used} & \textbf{Batch Size} & \textbf{Steps} \\ \hline
BioALBERT 1.0 & Base & PubMed & 4.3B & GCP TPU v3-8 & 1024 & 200K \\ \hline
BioALBERT 1.0 & Base & PubMed+PMC & 18.8B & GCP TPU v3-8 & 1024 & 470K \\ \hline
BioALBERT 1.1 & Large & PubMed & 4.3B & GCP TPU v3-8 & 256 & 200K \\ \hline
BioALBERT 1.1 & Large & PubMed+PMC & 18.8B & GCP TPU v3-8 & 256 & 470K \\ \hline
\end{tabular}%
}
\end{table}



Sentence embeddings are used for tokenization, and for that, we pre-processed the data as a sentence text. So every line in the input text document is a sentence, and an empty line separates every document. We set the maximum length of each sentence to 512 words. Shorter sentences were padded to make 512 whereas longer sentences were truncated. Learning rate of 0.00176 and number of warm-up steps as 3125 for all the pre-trained versions of the model, except for BioALBERT 1.1 (PubMed +PMC) where the learning rate has been re-adjusted to 0.00062 (by re-scaling the current learning rate with scaled $1/2^{1.5}$ times to avoid the problem of exploding loss.  Summary of parameters used in the training process is given in Table~\ref{parameters}.

We experimented with different settings and found out that both base and large models were successful with larger batch size on V3-8 TPU compute instances. We have used two different embedding sizes, i.e., 128 and 256. The 128 embedding size creates a base model with 12 Million parameters, whereas a large model has 16 Million parameters with 256 embedding size. With these combinations, we have presented a total of four models, given in Table~\ref{versions}.

\subsection{Fine-tuning of BioALBERT}




Fine-tuning of BioALBERT on BioNER task is presented in this section. BioNER involves annotating words in a sentence as named-entities.  The labelled datasets that were used for this task include four categories representing Disease, Species, Drug/Proteins, and Drugs/Chemicals. The objective is to train and make a prediction on the labels, which are the proper nouns within the domain area. More formally, given an input sentence  \(S = \{x_1,x_2,...,x_z\} \),  where $x_{i}$ is the $i$-th word and $z$ represents the length of the sentence. The goal of BioNER is to classify each word in $S$ and assign it to a corresponding label \(y \in Y\), where $Y$ is a predefined list of all possible label.


Fine-tuning is simple as compared to the pre-training, and the computational requirements are also not that significant. BioALBERT which takes less physical memory and improvised parameter sharing techniques. The BioNER fine-tuning is trained to learn the word embeddings using the sentence piece tokenization while the BioBERT model was based on the word piece embeddings. For each of the pre-trained models, we constructed a fine-tuning task by using the specific dataset. 

The model setup uses the weights of the pre-trained model that are created previously.  We have used  1e-5 learning rate, batch size as 32 and lower case texts and finally fine-tuned for 5336 steps. Pre-trained BioALBERT models were pre-trained using TPUv3-8. All of the hyper-parameters used are same as default ALBERT unless stated otherwise. All the tested datasets contain a list of words along with a label “B”, “I”, and “O” where “B” denotes the beginning of an entity, “I” stands for “inside” and is used for all words comprising the entity except the first one, and “O” means the absence of an entity. For our experiments, we used these datasets as-is and passed to it our pre-trained models for the downstream task. The adamw optimiser was used with evaluation checkpoint so that the model will be evaluated at different time intervals using the holdout development dataset to identify the best model for final predictions. The predictions were performed on the test datasets, and the performance is compared with baseline models that were established previously by calculating the F1 Score, Precision and Recall. The summary of all parameters used in fine-tuning is given in Table~\ref{finetuning}. 

\begin{table}[!tpb]
\centering
\caption{Summary of parameters used in fine-tuning}
\label{finetuning}
\begin{tabular}{|c|c|}
\hline
\multicolumn{2}{|c|}{\textbf{Summary   of All parameters used: (Fine Tuning)}} \\ \hline
\textbf{Name} & \textbf{BioNER} \\ \hline
Optimizer & adamw \\ \hline
Train Batch   Size & 32 \\ \hline
Eval Batch   Size & 16 \\ \hline
Save   Checkpoint & 200 \\ \hline
Max Seq Length & 512 \\ \hline
Learning Rate & 1.00E-05 \\ \hline
Training Steps & 5336 \\ \hline
Warmup   Steps & 320 \\ \hline
\end{tabular}%

\end{table}

\section{Experimental Analysis}\label{results}



In this section, we present the dataset used,  baselines and results to show the effectiveness of our model.

\subsection{\textbf{Datasets}}

Our model is evaluated on eight biomedical NER benchmark datasets which contain four types of entities and provided by Lee et al.~\cite{lee2019biobert}. Table \ref{stats_bioNER} shows the statistics of the datasets used. Below we briefly explained each dataset: 


    \begin{itemize}
        \item \textbf{BC5CDR}: The Bio-creative community challenge
for the chemical-disease relation extraction task
(BC5CDR) the corpus was made available in a Bio-creative
workshop~\cite{Li2016BioCreativeVC}. The two sub-tasks
of BC5CDR are identifying; (i) \textbf{chemical} and (ii) \textbf{disease}
entities from Medline abstracts. 




\item \textbf{BC4CHEMD}: This dataset is provided
by Bio-Creative community challenge IV for the development
and evaluation of tools for Chemical NER~\cite{CHEMDNERarticle}. BC4CHEMDNER was used
for the recognition of chemical compounds and drugs
from Pubmed abstracts. 



\item  \textbf{NCBI Disease}: To promote disease NER-system research, American National Institutes of Health released the NCBI disease corpus for disease NER-research. The NCBI disease corpus is large-scale and high-quality; it is based on the corpus released by Leaman et al.~\cite{10.5555/2772763.2772800}. 


\item \textbf{JNLPBA}: We also used the JNLPBA corpus in our experiments which are provided by Kim et al.~\cite{10.5555/1567594.1567610}. This corpus contains five entity types, including DNA, RNA, Cell Type, Cell Line and Protein. 


\item \textbf{BC2GM}: BC2GM is provided by Ando~\cite{Ando2007BioCreativeIG}, the state-of-the-art system in the Bio-Creative II gene mention recognition task is a semi-supervised learning method using alternating structure optimization.

\item \textbf{LINNAEUS}: The LINNAEUS corpus was provided by Gerner et al.~\cite{gerner2010linnaeus} which consists of 100 full-text documents from the PMC Open access document set which were randomly selected. All mentions of species terms were manually annotated and normalized to the NCBI taxonomy IDs of the intended species.


\item \textbf{Species-800}: Species-800, which is also known as S800~\cite{s800}, is a novel abstract-based manually annotated corpus. S800 comprises 800 PubMed abstracts in which organism mentions were identified and mapped to the corresponding NCBI taxonomy identifiers.


\end{itemize}

\begin{table}[!tpb]
\centering
\caption{Statistics of the BioNER
datasets}
\label{stats_bioNER}
\begin{tabular}{|c|c|c|}
\hline
\textbf{Entitity Type}        & \textbf{Dataset} & \textbf{\# of Annotations} \\ \hline
\multirow{2}{*}{Disease}      & NCBI Disease     & 6,881                      \\ \cline{2-3} 
                              & BC5CDR           & 12,694                     \\ \hline \hline
\multirow{2}{*}{Drug/Chem}    & BC5CDR           & 15,411                     \\ \cline{2-3} 
                              & BC4CHEMD         & 79,842                     \\ \hline \hline
\multirow{2}{*}{Drug/Protein} & BC2GN            & 20,703                     \\ \cline{2-3} 
                              & JNLPBA           & 35,460                     \\ \hline \hline
\multirow{2}{*}{SPECIES}      & LINNAEUS         & 4,077                      \\ \cline{2-3} 
                              & Species-800      & 3,708                      \\ \hline
\end{tabular}%

\end{table}

\begin{table*}[!htpb]
\centering
\caption{Comparison of performance in biomedical named entity recognition (BioNER) task.
}
\label{results2}
\resizebox{1.075\textwidth}{!}{%
\hspace{-15mm}\begin{tabular}{|c|c|c|c|c|c|c|c|c|c|c|c|}
\hline
\multirow{4}{*}{Type} & \multirow{4}{*}{Datasets} & \multirow{4}{*}{Metrics} & \multirow{4}{*}{SOTA } & \multirow{2}{*}{BioBERT v1.0} & \multirow{2}{*}{BioBERT v1.0} & \multirow{2}{*}{BioBERT v1.0} & \multirow{2}{*}{BioBERTv1.1} & \multirow{2}{*}{BioALBERT 1.0} & \multirow{2}{*}{BioALBERT 1.0} & \multirow{2}{*}{BioALBERT 1.1} & \multirow{2}{*}{BioALBERT 1.1} \\
 &  &  &  &  &  &  &  &  &  &  &  \\ \cline{5-12} 
 &  &  &  & (PubMed) & (PMC) & (PubMed+PMC) & (PubMed) & (PubMed) & (PubMed+PMC) & (PubMed) & (PubMed+PMC) \\ \cline{5-12} 
 &  &  &  & Base & Base & Base & Base & Base & Base & Large & Large \\ \hline
\multirow{6}{*}{Disease} & \multirow{3}{*}{NCBI Disease} & p & 88.30  & 86.76 & 86.16 & \underline{89.04} & 88.22 & \textbf{97.45} & 96.84 & 97.18 & 97.38 \\ \cline{3-12} 
 &  & R & 89.00 & 88.02 & 89.48 & 89.69 & \underline{91.25} & 94.39 & 94.40 & \textbf{97.18} & 94.37 \\ \cline{3-12} 
 &  & F & 88.60  & 87.38 & 87.79 & 89.36 & \underline{89.71}  & 95.89 & 95.61 & \textbf{97.18} & 95.85 \\ \cline{2-12} 
 & \multirow{3}{*}{BC5CDR} & p & \underline{89.61}  & 85.80 & 84.67 & 85.86 & 86.47 & \textbf{99.69} & 99.11 & 99.27 & 99.39 \\ \cline{3-12} 
 &  & R & 83.09  & \underline{86.60} & 85.87 & 87.27 & 87.84 & 95.72 & 96.17 & \textbf{96.33} & 95.85 \\ \cline{3-12} 
 &  & F & 86.23  & 86.2 & 85.27 & 86.56 & \underline{87.15} & 97.66 & 97.62 & \textbf{97.78} & 97.61 \\ \hline
\multirow{6}{*}{Drug/Chem} & \multirow{3}{*}{BC5CDR} & p & \underline{94.26}  & 92.52 & 92.46 & 93.27 & 93.68 & \textbf{99.99} & \textbf{99.99} & \textbf{99.99} & \textbf{99.99} \\ \cline{3-12} 
 &  & R & 92.38  & 92.76 & 92.63 & \underline{93.61} & 93.26 & 95.89 & \textbf{96.24} & 95.62 & 95.68 \\ \cline{3-12} 
 &  & F & 93.31  & 92.64 & 92.54 & 93.44 &\underline{93.47} & 97.9 & \textbf{98.08} & 97.76 & 97.79 \\ \cline{2-12} 
 & \multirow{3}{*}{BC4CHEMD} & p & 92.29  & 91.77 & 91.65 & 92.23 & \underline{92.80} & 97.76 & 97.71 & 97.71 & \textbf{97.88} \\ \cline{3-12} 
 &  & R & 90.01  & 90.77 & 90.30 & 90.61 & \underline{91.92} & 94.22 & \textbf{94.83} & \textbf{94.83} & 94.63 \\ \cline{3-12} 
 &  & F & 91.14  & 91.26 & 90.97 & 91.41 & \underline{92.36} & 95.96 & \textbf{96.25} & \textbf{96.25} & 96.23 \\ \hline
\multirow{6}{*}{Drug/Protein} & \multirow{3}{*}{BC2GM} & p & 81.81  & 81.72 & 82.86 & \underline{85.16} & 84.32 & 97.86 & 97.84 & \textbf{98.26} & 98.02 \\ \cline{3-12} 
 &  & R & 81.57  & 83.38 & 84.21 & 83.65 & \underline{85.12} & 94.87 & 94.27 & \textbf{95.72} & 94.70 \\ \cline{3-12} 
 &  & F & 81.69  & 82.54 & 83.53 & 84.40 & \underline{84.72} & 96.34 & 96.02 & \textbf{96.97} & 96.33 \\ \cline{2-12} 
 & \multirow{3}{*}{JNLPBA} & p & \underline{74.43}  & 71.11 & 71.17 & 72.68 & 72.24 & 85.14 & \textbf{85.60} & 86.23 & 85.56 \\ \cline{3-12} 
 &  & R & 83.22  & 83.11 & 82.76 & 83.21 & \underline{83.56} & 80.43 & 80.98 & \textbf{81.90} & 81.49 \\ \cline{3-12} 
 &  & F & 78.58  & 76.65 & 76.53 & \underline{77.59} & 77.49 & 82.72 & 83.22 & \textbf{84.01} & 83.53 \\ \hline
\multirow{6}{*}{Species} & \multirow{3}{*}{LINNAEUS} & p & 92.80  & 91.83 & 91.62 & \underline{ 93.84} & 90.77 & 99.95 & \textbf{99.98} & \textbf{99.98} & 99.92 \\ \cline{3-12} 
 &  & R & \underline{ 94.29}   & 84.72 & 85.48 & 86.11 & 85.83 & 99.47 & 99.44 & 99.48 & \textbf{99.55} \\ \cline{3-12} 
 &  & F & \underline{93.54}   & 88.13 & 88.45 & 89.81 & 88.24 & 99.71 & 99.72 & \textbf{99.73} & \textbf{99.73} \\ \cline{2-12} 
 & \multirow{3}{*}{Species-800} & p & \underline{74.34}   & 70.60 & 71.54 & 72.84 & 72.80 & \textbf{99.17} & 98.93 & 99.10 & 98.75 \\ \cline{3-12} 
 &  & R & 75.96   & 75.75 & 74.71 & \underline{77.97} & 75.36 & 98.34 & 98.04 & \textbf{98.95} & 98.69 \\ \cline{3-12} 
 &  & F & 74.98   & 73.08 & 73.09 & \underline{75.31} & 74.06 & 98.76 & 98.49 & \textbf{99.02} & 98.72 \\ \hline
\end{tabular}%
}
\\[10pt]
\caption*{\scriptsize Notes: Precision (P), Recall (R) and F1 (F) scores on each dataset are reported. The best scores are in \textbf{bold}, and the second best scores are \underline{underlined}. We list
the scores of the state-of-the-art (SOTA) models on different datasets as follows: scores of Xu et al.~\cite{Xu2019DocumentlevelAB} on NCBI Disease, scores of Sachan et al.~\cite{sachan2017effective} on
BC2GM, scores of Lou et al.~\cite{louarticle} on BC5CDR-disease, scores of Luo et al.~\cite{Luo2018AnAB} on
BC4CHEMD, scores of Yoon et al.~\cite{Yoon_2019} on BC5CDR-chemical and JNLPBA and scores of Giorgi and Bader~\cite{giorgi2018transfer} on LINNAEUS and Species-800}
\end{table*}

\subsection{Baselines}





To assess the performance of the proposed method, an exhaustive comparison is performed with several advanced SOTA methodologies along with their published results\footnote{The published results were acquired from respective original publication}. Our model is compared with the following methods.

Yoon et al.~\cite{Yoon_2019} introduced CollaboNet, which consists of multiple BiLSTM-CRF models, for BioNER. While existing models were only able to handle datasets with a single entity type, CollaboNet leverages multiple datasets and achieves the highest F1 scores. CollaboNet is built upon multiple single-task NER models (STMs) that send information to each other for more accurate predictions.



Lou et al.~\cite{louarticle} proposed a transition-based model for joint disease entity-recognition and normalization, based on transition-based structured prediction framework using structured perceptron with early-update training and beam-search decoding. In another study, Lou et al.~\cite{Luo2018AnAB} proposed a neural network approach, i.e. attention-based bidirectional Long Short-Term Memory with a conditional random field layer (Att-BiLSTM-CRF), to document level chemical NER. The method leverages document-level global information obtained by attention mechanism to enforce tagging consistency across multiple instances of the same token in a document. It achieves better performances with little feature engineering. 

Xu et al.~\cite{Xu2019DocumentlevelAB}  proposed a novel dictionary-based and document-level attention mechanism with a deep neural network NER method, named as DABLC. 
DABLC tags the consistency of multiple instances in a document at the document level and combines an external disease dictionary that is constructed with five disease resources containing a rich collection of disease entities. The authors adopted the efficient exact string matching method for dictionary matching; this method can effectively and accurately match the disease names.



Lee et al.~\cite{lee2019biobert} introduced BioBERT, which is a pre-trained language model for biomedical text mining. Authors showed that pre-training BERT on biomedical corpora is crucial in applying it to the biomedical domain. BioBERT outperforms previous models on biomedical text mining tasks such as NER, RE and QA. We compare BioALBERT with both BioBERT (v1.0 and V1.1) models and other SOTA models used in BioNER task. We selected those methods because they are the SOTA, and based on the conducted meta-analysis exhibit the highest performance among the techniques so far developed.

\subsection{Results}

\begin{figure*}[!htpb]
\centering
\includegraphics[width=1.00\linewidth]{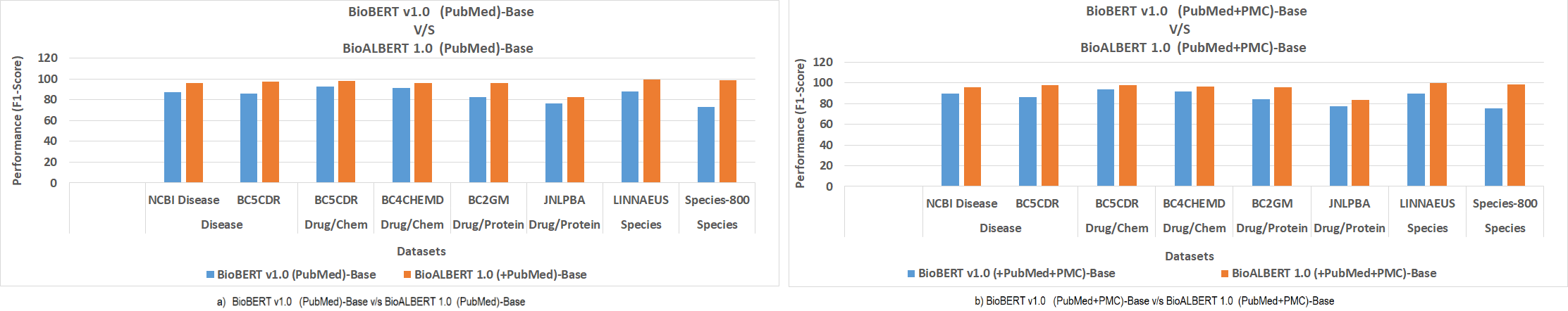}
\caption{\scriptsize Comparison of a) BioBERT v1.0 (Pubmed)-Base v/s BioALBERT 1.0 (Pubmed)-Base ; b)  BioBERT v1.0(Pubmed+PMC)-Base v/s BioALBERT 1.0(Pubmed+PMC)-Base}
\label{fig_sim33}
\end{figure*}

Table~\ref{results2} presents the performance of all the variants of BioALBERT and contrasts them to baseline methodologies. Our model outperforms all other methods on all eight datasets and entity types. For, (i) Disease-type datasets, BioALBERT improved the performance by 7.47\% and 10.63\% for NCBI-Disease and BC5CDR-Disease datasets respectively; (ii) Drug/Chem type datasets increase in performance by 4.61\% and 3.89\% for BC5CDR-Chem and BC4CHEMD datasets respectively; (iii) Gene/Protein type datasets increase in performance by 12.25\% and 6.42\% for BC2GM and JNLPBA datasets respectively and; (iv) Species type datasets increase in performance by 6.19\% and 23.71\% respectively is observed.

\begin{figure}[!htp]
\centering
\includegraphics[width=1\linewidth]{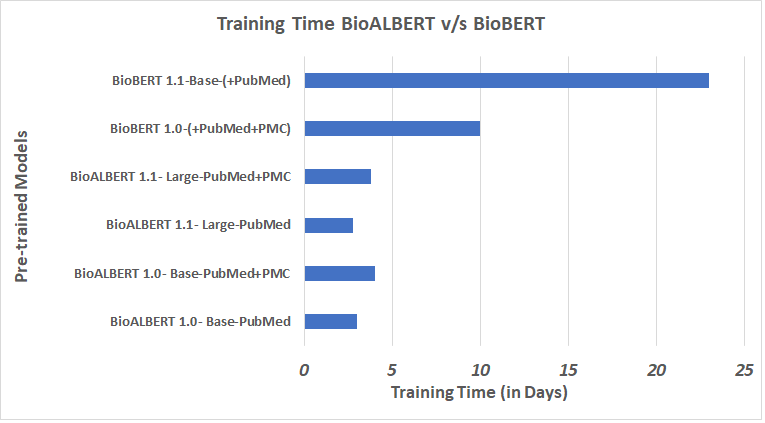}
\caption{\scriptsize Comparison of run-time statistics of BioALBERT v/s BioBERT}
\label{runstats}
\end{figure}

We have performed multiple comparisons to analyse the effectiveness of BioALBERT. Fig.~\ref{fig_sim33} presents, the performance comparison of the same versions (trained on same corpora and for the same number of steps) of both BioALBERT and BioBERT. We can see that in Fig.~(\ref{fig_sim33}a), we compared BioBERT v1.0-base model which is trained on PubMed with BioALBERT 1.0-base model trained on Pubmed and similarly in Fig.~(\ref{fig_sim33}b), we compared BioBERT v1.0-base model trained on Pubmed and PMC with BioALBER 1.0-base model trained on PubMed and PMC biomedical corpora. In both cases, BioALBERT outperformed BioBERT on all eight datasets. We also compared the training time of BioALBERT with BioBERT. We found that all models BioALBERT beat BioBERT with a considerable margin. The run time statistics of both pre-trained models are given in Fig.~\ref{runstats}.

\subsection{Discussions}

BioALBERT gives better performance and addresses the previously mentioned challenges in the biomedical domain. We attribute this to the BioALBERT built on top of the transformer-based language model that learns contextual relations between words (or subwords) in a text. As opposed to directional models, which read the text input sequentially (left-to-right or right-to-left), the transformer encoder reads the entire sequence of words at once. This characteristic allows the model to learn the context of a word based on all of its surroundings (left and right of the word) and address the issue of contextual representation.




Our model addresses the shortcomings of BERT based biomedical models. At first, BioALBERT uses cross-layer parameter sharing and reduces 110 million parameters of 12-layer BERT-base model to 31 million parameters while keeping the same number of layers and hidden units by learning parameters for the first block and reuse the block in the remaining 11 layers.  Secondly, our model uses the SOP, which takes two segments from the training corpus that appear consecutively and constructs a random pair of segments from different documents. This enables the model to learn about discourse-level coherence characteristics from a finer-grained distinction and leads to better learning representation in downstream tasks. Thirdly, our model uses factorized embedding parameterization in which a smaller size layer vocabulary and hidden layer to decompose the embedding matrix into two small matrices which reduce the number of parameters between vocabulary and the first hidden layer whereas in BERT based biomedical models embedding size is equal to the size of the hidden layer.  Furthermore, finally, our model is trained on massive biomedical corpora to be effective on BioNER to address the issue of the shift of word distribution from general domain corpora to biomedical corpora. All these, when combined, address all the issues associated with BioNER earlier. As our model offers a consistent improvement over all other methods for all tested datasets, we can conclude that it is a robust solution for BioNER. 

\begin{figure}[!htpb]
\centering
\includegraphics[width=1\linewidth]{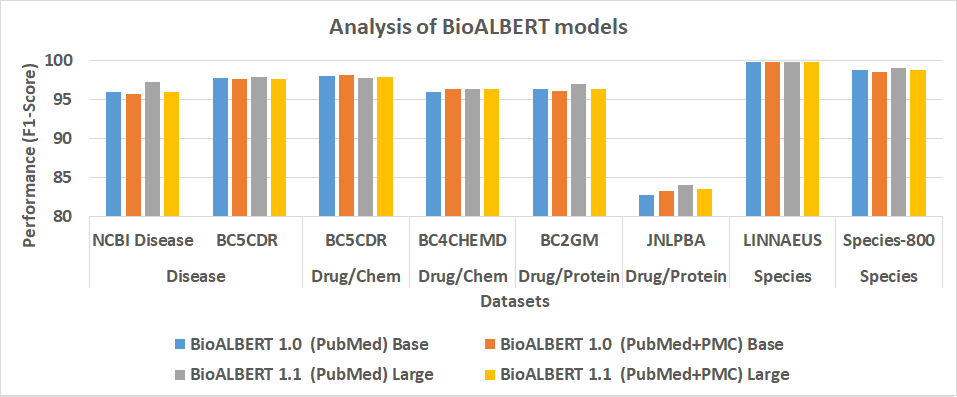}
\caption{Comparison of different variants BioALBERT models}
\label{biovaiarnt}
\end{figure}

To extend our analysis, we analysed the performance of different pre-trained models of BioALBERT. We found out that the performance of all BioALBERT models are almost equally good, but BioALBERT 1.1-large trained on PubMed works better than others (shown in Fig. \ref{biovaiarnt}). BioALBERT 1.1-large model, which is trained on PubMed with dup-factor as five performs better on most of the datasets. This shows that the relevance of duplication data in NLP tasks.

\section{Conclusion}\label{con}

In this study, we presented BioALBERT, which is a pre-trained language model for biomedical named entity recognition. We presented four different variants of BioALBERT models which are trained on huge biomedical corpora for a different number of steps. We showed that training ALBERT on biomedical corpora is a crucial step in applying it to BioNER. As future works, we plan to pre-trained other versions which include hybrid of general and biomedical corpora of ALBERT on biomedical corpora with more training steps and fine-tune on biomedical text mining task. We also plan to fine-tune BioALBERT on other text mining tasks to show the effectiveness of our model.


\begin{thebibliography}{10}

\bibitem{meystre2008extracting}
St{\'e}phane~M Meystre, Guergana~K Savova, Karin~C Kipper-Schuler, and John~F
  Hurdle.
\newblock Extracting information from textual documents in the electronic
  health record: a review of recent research.
\newblock {\em Yearbook of medical informatics}, 17(01):128--144, 2008.

\bibitem{maartensson2012health}
Lena M{\aa}rtensson and Gunnel Hensing.
\newblock Health literacy--a heterogeneous phenomenon: a literature review.
\newblock {\em Scandinavian journal of caring sciences}, 26(1):151--160, 2012.

\bibitem{vvarticle}
Sangrak Lim, Kyubum Lee, and Jaewoo Kang.
\newblock Drug drug interaction extraction from the literature using a
  recursive neural network.
\newblock {\em PLOS ONE}, 13:e0190926, 01 2018.

\bibitem{rosario-hearst-2004-classifying}
Barbara Rosario and Marti Hearst.
\newblock Classifying semantic relations in bioscience texts.
\newblock In {\em Proceedings of the 42nd Annual Meeting of the Association for
  Computational Linguistics ({ACL}-04)}, pages 430--437, Barcelona, Spain, July
  2004.

\bibitem{cohen2005survey}
Aaron~M Cohen and William~R Hersh.
\newblock A survey of current work in biomedical text mining.
\newblock {\em Briefings in bioinformatics}, 6(1):57--71, 2005.

\bibitem{yadav2019survey}
Vikas Yadav and Steven Bethard.
\newblock A survey on recent advances in named entity recognition from deep
  learning models, 2019.

\bibitem{peter}
Matthew~E. Peters, Mark Neumann, Mohit Iyyer, Matt Gardner, Christopher Clark,
  Kenton Lee, and Luke Zettlemoyer.
\newblock Deep contextualized word representations.
\newblock {\em CoRR}, abs/1802.05365, 2018.

\bibitem{devlin-etal-2019-bert}
Jacob Devlin, Ming-Wei Chang, Kenton Lee, and Kristina Toutanova.
\newblock {BERT}: Pre-training of deep bidirectional transformers for language
  understanding.
\newblock In {\em Proceedings of the 2019 Conference of the North {A}merican
  Chapter of the Association for Computational Linguistics: Human Language
  Technologies, Volume 1 (Long and Short Papers)}, pages 4171--4186,
  Minneapolis, Minnesota, June 2019. Association for Computational Linguistics.

\bibitem{lan2019albert}
Zhenzhong Lan, Mingda Chen, Sebastian Goodman, Kevin Gimpel, Piyush Sharma, and
  Radu Soricut.
\newblock Albert: A lite bert for self-supervised learning of language
  representations, 2019.

\bibitem{Pyysalo2013DistributionalSR}
Sampo Pyysalo, Filip Ginter, Hans Moen, Tapio Salakoski, and Sophia Ananiadou.
\newblock Distributional semantics resources for biomedical text processing.
\newblock 2013.

\bibitem{jin2019probing}
Qiao Jin, Bhuwan Dhingra, William~W. Cohen, and Xinghua Lu.
\newblock Probing biomedical embeddings from language models, 2019.

\bibitem{lee2019biobert}
Jinhyuk Lee, Wonjin Yoon, Sungdong Kim, Donghyeon Kim, Sunkyu Kim, Chan~Ho So,
  and Jaewoo Kang.
\newblock Biobert: a pre-trained biomedical language representation model for
  biomedical text mining, 2019.

\bibitem{mikolov2013distributed}
Tomas Mikolov, Ilya Sutskever, Kai Chen, Greg~S Corrado, and Jeff Dean.
\newblock Distributed representations of words and phrases and their
  compositionality.
\newblock In {\em Advances in neural information processing systems}, pages
  3111--9, 2013.

\bibitem{Chen2018BioSentVecCS}
Qingyu Chen, Yifan Peng, and Zhiyong Lu.
\newblock Biosentvec: creating sentence embeddings for biomedical texts.
\newblock {\em 2019 IEEE International Conference on Healthcare Informatics
  (ICHI)}, pages 1--5, 2018.

\bibitem{beltagy2019scibert}
Iz~Beltagy, Kyle Lo, and Arman Cohan.
\newblock Scibert: A pretrained language model for scientific text, 2019.

\bibitem{Zhu_Paschalidis_Tahmasebi_2018}
Henghui Zhu, Ioannis~C. Paschalidis, and Amir~M. Tahmasebi.
\newblock Clinical concept extraction with contextual word embedding.
\newblock {\em NIPS Machine Learning for Health Workshop}, Dec 2018.

\bibitem{Si_2019}
Yuqi Si, Jingqi Wang, Hua Xu, and Kirk Roberts.
\newblock Enhancing clinical concept extraction with contextual embeddings.
\newblock {\em Journal of the American Medical Informatics Association},
  26(11):1297--1304, Jul 2019.

\bibitem{peng2019transfer}
Yifan Peng, Shankai Yan, and Zhiyong Lu.
\newblock Transfer learning in biomedical natural language processing: An
  evaluation of bert and elmo on ten benchmarking datasets, 2019.

\bibitem{howard}
Jeremy Howard and Sebastian Ruder.
\newblock Universal language model fine-tuning for text classification.
\newblock In {\em Proceedings of the 56th Annual Meeting of the Association for
  Computational Linguistics (Volume 1: Long Papers)}, pages 328--339,
  Melbourne, Australia, July 2018. Association for Computational Linguistics.

\bibitem{yogatama2019learning}
Dani Yogatama, Cyprien de~Masson d'Autume, Jerome Connor, Tomas Kocisky, Mike
  Chrzanowski, Lingpeng Kong, Angeliki Lazaridou, Wang Ling, Lei Yu, Chris
  Dyer, et~al.
\newblock Learning and evaluating general linguistic intelligence.
\newblock {\em arXiv preprint arXiv:1901.11373}, 2019.

\bibitem{qiu2020pre}
Xipeng Qiu, Tianxiang Sun, Yige Xu, Yunfan Shao, Ning Dai, and Xuanjing Huang.
\newblock Pre-trained models for natural language processing: A survey.
\newblock {\em arXiv preprint arXiv:2003.08271}, 2020.

\bibitem{liu2016recurrent}
Pengfei Liu, Xipeng Qiu, and Xuanjing Huang.
\newblock Recurrent neural network for text classification with multi-task
  learning.
\newblock {\em arXiv preprint arXiv:1605.05101}, 2016.

\bibitem{liu2019multi}
Xiaodong Liu, Pengcheng He, Weizhu Chen, and Jianfeng Gao.
\newblock Multi-task deep neural networks for natural language understanding.
\newblock {\em arXiv preprint arXiv:1901.11504}, 2019.

\bibitem{Li2016BioCreativeVC}
Jiao Li, Yueping Sun, Robin~J. Johnson, Daniela Sciaky, Chih-Hsuan Wei, Robert
  Leaman, Allan~Peter Davis, Carolyn~J. Mattingly, Thomas~C. Wiegers, and
  Zhiyong Lu.
\newblock Biocreative v cdr task corpus: a resource for chemical disease
  relation extraction.
\newblock {\em Database : the journal of biological databases and curation},
  2016, 2016.

\bibitem{CHEMDNERarticle}
Martin Krallinger, Obdulia Rabal, Florian Leitner, Miguel Vazquez, David
  Salgado, Zhiyong lu, Robert Leaman, Yanan Lu, Donghong Ji, Daniel Lowe, Roger
  Sayle, Riza Batista-Navarro, Rafal Rak, Torsten Huber, Tim Rocktäschel,
  Sérgio Matos, David Campos, Buzhou Tang, Hua Xu, and Alfonso Valencia.
\newblock The chemdner corpus of chemicals and drugs and its annotation
  principles.
\newblock {\em Journal of Cheminformatics}, 7:S2, 03 2015.

\bibitem{10.5555/2772763.2772800}
Rezarta~Islamaj Doundefinedan, Robert Leaman, and Zhiyong Lu.
\newblock Ncbi disease corpus.
\newblock {\em J. of Biomedical Informatics}, 47(C):1â€“10, February 2014.

\bibitem{10.5555/1567594.1567610}
Jin-Dong Kim, Tomoko Ohta, Yoshimasa Tsuruoka, Yuka Tateisi, and Nigel Collier.
\newblock Introduction to the bio-entity recognition task at jnlpba.
\newblock In {\em Proceedings of the International Joint Workshop on Natural
  Language Processing in Biomedicine and Its Applications}, JNLPBA â€™04, page
  70â€“75, USA, 2004. Association for Computational Linguistics.

\bibitem{Ando2007BioCreativeIG}
Rie~Kubota Ando.
\newblock Biocreative ii gene mention tagging system at ibm watson.
\newblock 2007.

\bibitem{gerner2010linnaeus}
Martin Gerner, Goran Nenadic, and Casey~M Bergman.
\newblock Linnaeus: a species name identification system for biomedical
  literature.
\newblock {\em BMC bioinformatics}, 11(1):85, 2010.

\bibitem{s800}
Evangelos Pafilis, Sune~P. Frankild, Lucia Fanini, Sarah Faulwetter, Christina
  Pavloudi, Aikaterini Vasileiadou, Christos Arvanitidis, and Lars~Juhl Jensen.
\newblock The species and organisms resources for fast and accurate
  identification of taxonomic names in text.
\newblock {\em PLOS ONE}, 8(6):1--6, 06 2013.

\bibitem{Xu2019DocumentlevelAB}
Kai Xu, Zhenguo Yang, Peipei Kang, Qi~Wang, and Wenyin Liu.
\newblock Document-level attention-based bilstm-crf incorporating disease
  dictionary for disease named entity recognition.
\newblock {\em Computers in biology and medicine}, 108:122--132, 2019.

\bibitem{sachan2017effective}
Devendra~Singh Sachan, Pengtao Xie, Mrinmaya Sachan, and Eric~P Xing.
\newblock Effective use of bidirectional language modeling for transfer
  learning in biomedical named entity recognition, 2017.

\bibitem{louarticle}
Yinxia Lou, Yue Zhang, Tao Qian, Fei Li, Shufeng Xiong, and Donghong Ji.
\newblock A transition-based joint model for disease named entity recognition
  and normalization.
\newblock {\em Bioinformatics (Oxford, England)}, 33, 03 2017.

\bibitem{Luo2018AnAB}
Ling Luo, Zhihao Yang, Pei Yang, Yin Zhang, Lei Wang, Hongfei Lin, and Jian
  Wang.
\newblock An attention based bilstm crf approach to document level
  chemical named entity recognition.
\newblock {\em Bioinformatics}, 34:1381â€“1388, 2018.

\bibitem{Yoon_2019}
Wonjin Yoon, Chan~Ho So, Jinhyuk Lee, and Jaewoo Kang.
\newblock Collabonet: collaboration of deep neural networks for biomedical
  named entity recognition.
\newblock {\em BMC Bioinformatics}, 20(S10), May 2019.

\bibitem{giorgi2018transfer}
John~M Giorgi and Gary~D Bader.
\newblock Transfer learning for biomedical named entity recognition with neural
  networks.
\newblock {\em Bioinformatics}, 34(23):4087--4094, 2018.

\end{thebibliography}
\end{document}